\documentclass[twocol,final]{ametsocV6.1}
\usepackage{changes}      

\definechangesauthor[name=Riwal, color=bl
ue]{Riwal}
\definechangesauthor[name=Camille, color=purple]{Camille}
\definechangesauthor[name=Ganglin, color=red]{Ganglin}

\title{Quantile Regression, Variational Autoencoders, and Diffusion Models for Uncertainty Quantification: A Spatial Analysis of Sub-seasonal Wind Speed Prediction}
\thanks{This Work has been submitted to Monthly Weather Review. Copyright in this Work may be transferred without further notice.}

\authors{
Ganglin Tian,\aff{a}\correspondingauthor{ganglin.tian@lmd.ipsl.fr}
Anastase Alexandre Charantonis,\aff{a,b}
Camille Le Coz,\aff{a}
Alexis Tantet,\aff{a}
and Riwal Plougonven\aff{a}
}

\affiliation{\aff{a}{LMD/IPSL, École Polytechnique, Institut Polytechnique de Paris, ENS,Université PSL, Sorbonne Université, CNRS, Palaiseau, 91120, France}\\
\aff{b}{INRIA, Paris, France}\\
}

\abstract{This study aims to improve the spatial representation of uncertainties when regressing surface wind speeds from large-scale atmospheric predictors for sub-seasonal forecasting. Sub-seasonal forecasting often relies on large-scale atmospheric predictors such as 500 hPa geopotential height (Z500), which exhibit higher predictability than surface variables and can be downscaled to obtain more localised information. Previous work by Tian et al. (2024) demonstrated that stochastic perturbations based on model residuals can improve ensemble dispersion representation in statistical downscaling frameworks, but this method fails to represent spatial correlations and physical consistency adequately. More sophisticated approaches are needed to capture the complex relationships between large-scale predictors and local-scale predictands while maintaining physical consistency. Probabilistic deep learning models offer promising solutions for capturing complex spatial dependencies. This study evaluates three probabilistic methods with distinct uncertainty quantification mechanisms: Quantile Regression Neural Network that directly models distribution quantiles, Variational Autoencoders that leverage latent space sampling, and Diffusion Models that utilise iterative denoising. \replaced[id=Ganglin]{These models are trained on ERA5 reanalysis data and applied to ECMWF sub-seasonal hindcasts to regress probabilistic wind speed ensembles.}{These models are implemented in a two-stage framework: first trained on ERA5 reanalysis data to establish spatial uncertainty representation from 500 hPa geopotential height to surface wind speeds, then applied to ECMWF hindcasts to generate wind speed ensembles.} Our results show that probabilistic downscaling approaches provide more realistic spatial uncertainty representations compared to simpler stochastic methods, with each probabilistic model offering different strengths in terms of ensemble dispersion, deterministic skill, and physical consistency. These findings establish probabilistic downscaling as an effective enhancement to operational sub-seasonal wind forecasts for renewable energy planning and risk assessment.}

\begin{document}

\maketitle

\section{Introduction}\label{secIntro}
As the penetration of wind energy in the global energy system increases and annual capacity additions continue to rise, sub-seasonal wind speed forecasts have become increasingly important for optimising turbine maintenance, resource allocation, and grid integration \citep{luzia2022evaluating, chen2019multifactor,tawn2022subseasonal,cassola2012wind,chang2014literature,white2017potential}. 

Statistical downscaling emerges as an essential bridge between large-scale forecasts and local wind conditions, offering a computationally efficient method for extracting predictability from large-scale circulations into surface wind speed forecasts \citep{tian2025improving,alonzo2017modelling,goutham2023statistical}. However, statistical downscaling methods face a fundamental challenge: they typically produce fields that are too smooth, restricting them to only the portion of the flow that can be estimated from predictors. As a result, they do not appropriately represent the full uncertainty in the system \citep{orth2014using,tian2025improving}. This limitation is most evident when these statistical downscaling methods are applied to dynamic ensemble forecasts, where the predicted spread under-estimates atmospheric uncertainty, a phenomenon known as the ``under-dispersion problem'' \citep{wilks_statistical_2019,leutbecher2019ensemble}. This under-dispersion occurs because deterministic downscaling models only capture the conditional mean of the predictand given predictors, neglecting the conditional variance that represents the inherent uncertainty \citep{von1999use}.

Statistical downscaling must also address the inherent spatial correlation structures within meteorological fields. Wind speed fields exhibit significant spatial dependencies, with studies demonstrating substantial correlations extending up to hundreds of kilometres between sites \citep{chen2019multifactor,hill2011application}. These correlations reflect the scales of the meteorological features driving the boundary layer winds: fronts at the meso-scale, up to storms on the synoptic scale \citep{holton2013introduction}. These spatial correlations are important for renewable energy applications, as they directly impact power system operations through their representation of regional wind patterns \citep{sperati2017gridded}. 
For instance, concurrent production peaks in correlated regions can lead to grid overload \citep{sperati2017gridded}, while skillful spatial forecasts are essential for transmission system operators managing reserve estimation and grid stability \citep{davo2016post}.

Our previous work \citep{tian2025improving} addressed the under-dispersion issue through stochastic perturbations, adding Gaussian noise derived from statistical model residuals to each grid point independently. While this method contributes to solving the under-dispersion as evaluated by skill scores calculated at each point, it relies on two assumptions that the uncertainty follows a Gaussian distribution and that spatial points are independent. These assumptions oversimplify ensemble uncertainty, particularly in terms of spatial coherence.

Recent advances in probabilistic modelling offer potential solutions to these limitations. Modern approaches such as quantile regression, variational autoencoders (VAEs), and diffusion models provide sophisticated frameworks for uncertainty quantification. Quantile regression enables direct modelling of the conditional distribution without assuming a specific distribution, making it particularly effective for capturing non-Gaussian uncertainty structures in wind speed. 
\citet{schulz2022machine} have demonstrated that quantile regression outperforms traditional methods in postprocessing wind speed and temperature forecasts, better capturing the variance of variables across different conditions and skewness of forecast distribution. VAEs have shown promise in wind speed forecasting by learning low-dimensional latent space representations that can encode complex spatial dependencies while generating spatially consistent ensemble members with plausible patterns. VAEs have demonstrated their ability to generate diverse samples that reflect system uncertainties in wind speed prediction \citep{zhong2024fuxia}. 
Diffusion models, through their iterative denoising process, can generate diverse yet coherent wind speed fields that preserve the spatial correlations and statistical properties of target fields. \citet{price2025probabilistic} and \citet{zhong2024fuxib} have demonstrated effectiveness in handling signal noise and improving prediction accuracy, particularly in extreme weather conditions. A key consideration for both VAEs and diffusion models is their ability to maintain atmospheric physical realism beyond mere statistical reproduction of historical patterns. These methods move beyond simple Gaussian assumptions and grid-wise independence, potentially addressing both the under-dispersion issue and the need for spatial coherence in ensemble forecasts. However, most existing applications of these methods have focused on short-term or medium-range forecasting horizons \citep{dong2023short,price2025probabilistic,lam2022graphcast,chen2025learning,zhong2024fuxib,zhong2024fuxia}. 
While these studies demonstrate the potential of probabilistic models in wind speed prediction, the effectiveness of these methods in capturing and representing spatial uncertainty at sub-seasonal time scales remains unexplored, particularly in statistical downscaling applications where the translation of large-scale predictors to local-scale wind fields introduces additional complexity and uncertainty. 

This study examines how these statistical methods capture spatial correlation structures in sub-seasonal wind speed ensembles\deleted[id=Ganglin]{ across different forecast lead times, advancing beyond the stochastic perturbation approach developed by 
}. \added[id=Ganglin]{Specifically, we aim to: (1) systematically compare how stochastic perturbations, quantile regression, variational autoencoders, and diffusion models preserve multi-scale spatial dependencies, (2) evaluate whether grid-wise metrics adequately assess the spatial coherence of probabilistic forecasts. }This study is structured as follows: 
\replaced[id=Ganglin]{
    Section \ref{secMethodology} explains our methodology, outlining the framework for uncertainty quantification and describing the statistical downscaling models, their architectures, and uncertainty representation mechanisms. Section \ref{secData} describes the datasets, including ERA5 reanalysis for training and ECMWF sub-seasonal hindcasts for evaluation, as well as preprocessing steps and evaluation metrics. Section \ref{secResults_HC} analyses model performance on sub-seasonal hindcasts, exploring how forecast skill evolves with lead time through spatially averaged scores, spatial distribution maps, EOF reconstruction, and spectrum analysis. Section \ref{secDiscussion} discusses our findings in relation to model mechanisms, practical implications, and limitations, and proposes directions for further improvement.
}{Section \ref{secMethodology} explains our methodology, first outlining our framework for uncertainty quantification, then describing the statistical downscaling models, their architectures, and uncertainty representation mechanisms. Section \ref{secData} describes the datasets used in this study, including ERA5 reanalysis and ECMWF sub-seasonal forecast ensembles, as well as preprocessing steps and definitions of evaluation metrics. \deleted[id=Ganglin]{Section \ref{secResults_RA} first presents the results on reanalysis, including spatially averaged skills, spatial distribution maps, principal component reconstruction analysis, and spectrum analysis. Then, }Section \ref{secResults_HC} analyses the skill of models on a sub-seasonal time scale, exploring how forecast skill evolves with forecast lead time and the relative advantages of different uncertainty representation methods in handling ensemble uncertainty. Section \ref{secDiscussion} provides a discussion of our approach and its limitations, trying to relate the results to the mechanisms of the models used, and proposes directions for further improvement. Finally, Section \ref{secConclusion} summarises the research findings and main contributions. }

\section{Methodology}\label{secMethodology}
\subsection{Framework}
This study extends our previous work \citep{tian2025improving} by introducing spatial uncertainty quantification methods within the same statistical downscaling framework. \deleted[id=Ganglin]{The proposed framework consists of two key stages: model training and validation on reanalysis, and prediction on ensembles. }
\replaced[id=Ganglin]{While \citet{tian2025improving} addressed ensemble under-dispersion through stochastic perturbations, the grid-independent application of these perturbations destroys spatial correlations. This study evaluates three advanced probabilistic methods that better preserve spatial structure: Quantile Regression Neural Network (QNN), Variational Autoencoder-based Neural Network (VNN), and Diffusion-based Neural Network (DNN). }{However, while 
addressed the under-dispersion of statistically downscaled ensembles by leveraging the predictability of large-scale atmospheric circulation patterns (500 hPa geopotential height, Z500) to improve surface wind speed forecasts, their grid-wise stochastic perturbation approach actually destroys small-scale spatial correlations. Since the random perturbations are independent for each grid point and time step, spatial and temporal consistency are destroyed, potentially compromising the physical plausibility of the generated ensembles. }

\replaced[id=Ganglin]{All models are trained on ERA5 reanalysis to learn a probabilistic mapping from 500 hPa geopotential height ($\mathbf{X}$) to surface wind speed ($\mathbf{Y}$). We follow the data preprocessing procedures from \citet{tian2025improving}, including temporal and spatial aggregation and standardisation. Each model optimises specific loss functions detailed in Section \ref{secMethodology}\ref{subsecModels}. The trained models are then applied to ECMWF ensemble hindcasts for evaluation. }{To address these limitations, this study builds upon the original two-stage framework in 
while incorporating three advanced probabilistic modelling methods to better quantify regression uncertainty: Neural Network-based Quantile regression (QNN), Variational autoencoders (VNN), and Diffusion models (DNN), with the Stochastic Neural Network (SNN) in 
as a baseline for comparison. During the training phase on reanalysis, each of these models establishes a probabilistic mapping from a large-scale variable $\mathbf{X}$ to a surface-scale variable $\mathbf{Y}$. For data preprocessing, we followed the same procedures as in 
, including temporal and spatial resolution aggregation as well as standardisation (see appendix in 
for details). Each model optimises its parameters by minimising a specific loss function during training (Section \ref{secMethodology}\ref{subsecModels}). }
\replaced[id=Ganglin]{For ensemble forecasting, each trained model processes M-member ECMWF ensembles $\mathbf{X}^M$ to generate probabilistic forecasts. QNN produces $P=10$ quantiles per input member, while SNN, VNN, and DNN generate $P=20$ independent samples, yielding $P \times M$ total ensemble members for evaluation.
}{In the ensemble forecasting stage, the trained models are applied to M-member forecast ensembles $\mathbf{X}^M$. For the $m$-th input member $\mathbf{X}^m$, the four models employ distinct strategies to generate probabilistic forecasts (detailed in Section \ref{secMethodology}\ref{subsecModels}). 
To ensure fairness in comparison, we generated $P$ outputs for each input member: $P=10$ equally spaced quantiles (from 0.05 to 0.95) for QNN, and $P=20$ independent samples from SNN, VNN, and DNN. Consequently, for $M$ input members, the final output consisted of $P \times M$ ensemble members. }

\subsection{Models}\label{subsecModels}
\begin{figure*}[htb!]
    \centering
    \includegraphics[width=0.9\linewidth]{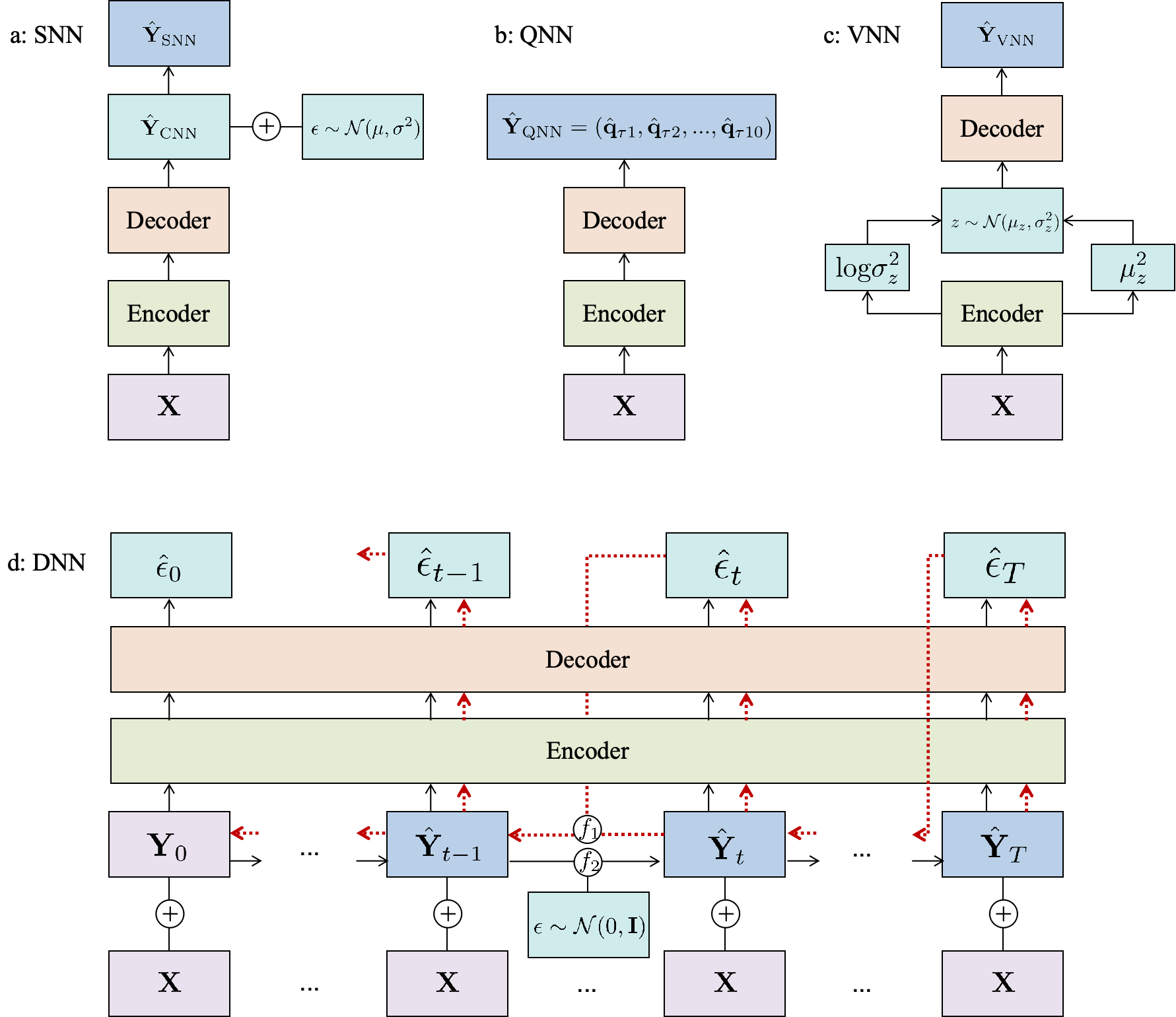}
    \caption{Architectures of four uncertainty quantification neural networks. (a) Stochastic Neural Network (SNN): SNN incorporates uncertainty by adding gaussian noise $\mathcal{N}(\mu, \sigma^2)$ to deterministic CNN outputs; (b) Quantile Regression Neural Network (QNN): QNN predicts ten specific quantiles ($(\hat{q}_{\tau_1}, \hat{q}_{\tau_2}, ..., \hat{q}_{\tau_{10}})$) simultaneously to characterise conditional distributions; (c) Variational Autoencoder-based Neural Network (VNN): VNN encodes inputs into a latent probability distribution by outputting mean $\mu_{z}$ and log-variance $\log\sigma^2_{z}$, then samples from $\mathcal{N}(\mu_{z}, \sigma^2_{z})$ to generate diverse outputs; (d) Diffusion-based Neural Network (DNN): DNN implements a bidirectional diffusion framework with two complementary processes: a forward process ($f_2$) that progressively adds noise to the clean target field $\mathbf{Y}_0$, and a reverse denoising process ($f_1$) that iteratively removes noise starting from a pure noise state $\hat{\mathbf{Y}}_T \sim \mathcal{N}(0,\mathbf{I})$. During inference, the model predicts noise components $\hat{\boldsymbol{\epsilon}}_t(\hat{\mathbf{Y}}_t,t,\mathbf{X})$ conditioned on both the noisy state and input $\mathbf{X}$ at each timestep, gradually transforming random noise into physically consistent predictions $\hat{\mathbf{Y}}_0$.}
    \label{fig:four_models}
\end{figure*}

Building upon the CNN architecture established by \citet{tian2025improving}, all probabilistic models employ SmaAt-UNet \citep{trebing2021smaat} backbone for consistent feature extraction capabilities. This ensures that performance differences can be attributed primarily to the uncertainty quantification mechanisms rather than variations in underlying spatial feature representation. Each model incorporates specific architectural modifications to implement its particular approach to uncertainty representation. Figure \ref{fig:four_models} illustrates the four uncertainty representation approaches investigated in this study. Each method represents a distinct theoretical framework for representing conditional distributions, allowing us to systematically evaluate how different uncertainty quantification strategies compare to the established stochastic perturbation approach. For a detailed overview of SmaAt-UNet structure, please refer to the appendix in \citet{tian2025improving}. Implementation details, including model configuration, hyperparameter optimisation, and training procedures, are provided in Appendix \ref{appendix:implementation}. 

\subsubsection{Quantile regression-based neural network}

Quantile regression extends beyond the conditional expectation $E(\mathbf{Y}|\mathbf{X})$ estimated by traditional regression methods (such as the multiple linear regression and CNN approaches described in \citet{tian2025improving}) to directly characterise conditional distributions. Rather than assuming specific distributional forms, this approach estimates the quantiles of the target $\mathbf{Y}$ conditioned on the input $\mathbf{X}$, denoted as $\hat{q_{\tau}}(\mathbf{Y}|\mathbf{X})$ at specified quantile levels $\tau \in (0,1)$, enabling the detection of distributional features such as the asymmetry that are masked by expectation-based methods. 

Building upon the CNN architecture established in \citet{tian2025improving}, QNN in Figure \ref{fig:four_models}b modifies the final convolutional layer of the SmaAt-UNet from single-channel to 10-channel output. Each output channel corresponds to a specific quantile level $\tau_p \in \{5\%, 15\%,\dots, 95\%\}$ for $p=1,2,...,10$, enabling simultaneous prediction of all quantiles $\hat{q_{\tau}}$ while sharing feature extraction weights across the encoder-decoder architecture. This 10\% interval selection, leading to $P=10$, provides balanced distributional coverage while maintaining computational efficiency. 

QNN is optimised using the loss function: 
\begin{equation}
    \mathcal{L}_{QNN} = \mathbb{E}_{n,g,p}[\rho_{\tau_p}(\mathbf{Y}_{n,g} - \hat{\mathbf{Y}}_{n,g}^{\tau_p})]
\end{equation}

where the expectation is taken over all training samples $n$, grid points $g$, and quantile levels $p$. In practice, this expectation is approximated by the empirical average: $\mathcal{L}_{QNN} = \frac{1}{N \cdot G \cdot 10}\sum_{n=1}^N \sum_{g=1}^G \sum_{p=1}^{10} \rho_{\tau_p}(\mathbf{Y}_{n,g} - \hat{\mathbf{Y}}_{n,g}^{\tau_p})$, where $N$ is the number of training samples, $G$ is the number of grid points, and 10 represents the number of quantile levels. Here, $\mathbf{Y}_{n,g}$ is the target value for sample $n$ at grid point $g$, $\hat{\mathbf{Y}}_{n,g}^{\tau_p}$ is the predicted quantile for a given quantile level $\tau_p$, representing the neural network's estimate of $\hat{q_{\tau_p}}(\mathbf{Y}|\mathbf{X})$, and $\rho_\tau(u) = u(\tau - \mathbb{I}(u < 0))$ is the quantile loss function, $\mathbb{I}(\cdot)$ is the indicator function. Unlike the MSE optimisation used in traditional regression, this asymmetric loss function assigns different penalties to positive and negative errors. Errors below the target are weighted by $\tau_p$, while errors above are weighted by $(1-\tau_p)$. By minimising this expected loss across multiple quantile levels, we obtain a comprehensive description of the conditional distribution. 

For ensemble forecasting, QNN processes each input ensemble member $\mathbf{X}^m$ to generate ten quantile predictions $\{\hat{\mathbf{Y}}^{\tau_p,m}\}_{p=1}^{10}$. For an input ensemble with $M$ members, this produces a $10 \times M$-member output ensemble. Rather than maintaining strict quantile ordering across ensemble members, these predictions are treated as exchangeable ensemble members that collectively represent the predictive distribution. This approach prioritises distributional coverage over formal quantile constraints, providing robust uncertainty representation while avoiding quantile crossing issues that can arise in multi-member ensemble contexts. 

\subsubsection{Variational autoencoder-based neural network}

VAEs provide an alternative approach to uncertainty quantification through latent space sampling. Rather than the direct input-output mapping employed in the CNN approach from \citet{tian2025improving}, VNN decomposes the prediction task into encoding and decoding stages. The encoder learns to approximate the posterior distribution $q(z|\mathbf{X},\mathbf{Y})$ of latent features $z$ conditioned on both input predictors and target variables, while the decoder models the conditional distribution $p(\hat{\mathbf{Y}}|z,\mathbf{X})$ for reconstruction given the latent representation and input conditions. 

As illustrated in Figure \ref{fig:four_models}c, the VNN architecture modifies the CNN backbone by introducing a bottleneck encoding stage that parameterises latent distributions rather than deterministic features. The encoder outputs mean $\mu_{z}$ and log-variance $\log\sigma^2_{z}$ vectors representing the parameters of the latent space distribution. During both training and inference, we employ the reparameterisation trick \citep{kingma2013auto} to enable gradient flow: $z = \mu_{z} + \sigma_{z} \odot \mathcal{N}(0, I)$, where $\odot$ denotes element-wise multiplication. For a given input, performing $P$ independent sampling operations from this learned latent distribution $\mathcal{N}(\mu_{z}, \sigma^2_{z})$ generates $P$ distinct latent representations, which are then decoded to produce diverse outputs $\hat{\mathbf{Y}}^{p}$ where $p=1,2,...,P$. This sampling-based diversity mechanism enables uncertainty quantification by capturing the variability inherent in the latent space. 

To optimise both regression accuracy and latent space regularisation, we implement a composite loss function: 
\begin{equation}
    \mathcal{L}_{VNN} = \mathbb{E}_{n,g}[(\mathbf{Y}_{n,g} - \hat{\mathbf{Y}}_{n,g})^2] + \beta \mathbb{E}_{n}[\mathcal{L}_{KL}]
\end{equation}

where the first expectation is taken over all training samples $n$ and grid points $g$, representing the reconstruction loss between predicted outputs $\hat{\mathbf{Y}}_{n,g}$ and target values $\mathbf{Y}_{n,g}$. The second expectation is taken over training samples, where the Kullback–Leibler (KL) divergence term for samples is: 
\begin{equation}
    \mathcal{L}_{KL} = -\frac{1}{2}\sum_{i=1}^{d_z}(1 + \log\sigma_{z,i}^{2} - \mu_{z,i}^{2} - \sigma_{z,i}^{2})
\end{equation}
This $\beta$-VAE formulation enables control over the information bottleneck, with higher $\beta$ values enforcing stronger regularisation but potentially reducing reconstruction fidelity. In practice, both expectations are approximated by empirical averages over the training dataset. 

For ensemble forecasting, each input member $\mathbf{X}^m$ undergoes $P$ independent sampling operations in the latent space, generating $P$ distinct predictions that reflect both the learned uncertainty structure and the stochastic sampling process. This approach combines the ensemble uncertainty from the original meteorological ensemble with the regression uncertainty captured by the VAE framework. For an input ensemble with $M$ members, this produces $P \times M$ output ensemble members $\hat{\mathbf{Y}}^{P \times M}$, providing comprehensive uncertainty representation. 

\subsubsection{Diffusion-based neural network}\label{subsecDiffusion}

Diffusion models establish a reversible mapping between data distributions at different noise levels through two complementary processes: a forward process that gradually adds noise to data until reaching a Gaussian distribution, and a reverse process that systematically removes noise to reconstruct the original data. Unlike QNN's direct quantile prediction or VNN's latent space sampling, DNN reframes the prediction task as progressive noise removal conditioned on input atmospheric patterns.

The diffusion framework, illustrated in Figure \ref{fig:four_models}d, consists of two distinct processes operating on the target field $\mathbf{Y}_0$ ($\mathbf{Y}_0$ is $\mathbf{Y}$, where subscript 0 indicates the initial clean state). We define $\hat{\mathbf{Y}}_t$ as the noisy version of the field at diffusion step $t$, with $\boldsymbol{\epsilon} \sim \mathcal{N}(0, \mathbf{I})$ representing gaussian noise. 

The forward diffusion process (denoted as $f_2$ in Figure \ref{fig:four_models}d) progressively adds noise over $T$ time steps: 
\begin{equation}
 f_2: \hat{\mathbf{Y}}_{t} = \sqrt{\alpha_t}\mathbf{Y}_0 + \sqrt{1-\alpha_t}\boldsymbol{\epsilon}
\end{equation}

At each step $t$, $\hat{\mathbf{Y}}_t$ follows a gaussian distribution with mean $\sqrt{\alpha_t} \mathbf{Y}_0$ and variance $(1 - \alpha_t)\mathbf{I}$. The coefficient $\alpha_t = \prod_{s=1}^t (1 - \beta_s)$ represents the cumulative noise schedule, where $\beta_t$ increases linearly from $\beta_{\text{start}}$ to $\beta_{\text{end}}$ over $T$ steps:
\begin{equation}
\beta_t = \beta_{\text{start}} + (t-1) \cdot \frac{\beta_{\text{end}} - \beta_{\text{start}}}{T-1}
\end{equation}

The reverse denoising process (denoted as $f_1$ in Figure \ref{fig:four_models}d) iteratively removes noise to generate predictions:
\begin{equation}
 f_1: \hat{\mathbf{Y}}_{t-1} = \frac{1}{\sqrt{1 - \beta_t}}\left(\hat{\mathbf{Y}}_t - \frac{\beta_t}{\sqrt{1-\alpha_t}} \hat{\boldsymbol{\epsilon}}_t(\hat{\mathbf{Y}}_t,t,\mathbf{X})\right) + \sqrt{\beta_t} \boldsymbol{\epsilon}
\end{equation},
where $\hat{\boldsymbol{\epsilon}}_t(\hat{\mathbf{Y}}_t,t,\mathbf{X})$ is the model's predicted noise component at step $t$ given input $\mathbf{X}$. 

During training, the model learns to predict the noise component $\hat{\boldsymbol{\epsilon}}_t$ added at each diffusion step. The loss function optimises noise prediction accuracy across all diffusion steps:
\begin{equation}
\mathcal{L}_{DNN} = \mathbb{E}_{t, n}[(\boldsymbol{\epsilon} - \hat{\boldsymbol{\epsilon}}_{t, n}(\hat{\mathbf{Y}}_t,t,\mathbf{X}))^2]
\end{equation}

The expectation is taken over uniformly sampled diffusion time steps and training samples. This formulation trains the model to understand what noise was added at each step of the diffusion process. 

For inference, we start with pure noise $\hat{\mathbf{Y}}_T \sim \mathcal{N}(0,\mathbf{I})$ and progressively denoise it through the reverse process. We set the hyperparameters $\beta_{\text{start}}$ and $\beta_{\text{end}}$ (detailed in Appendix \ref{appendix:implementation}) to ensure a gradual noise injection during training, with $\beta_{\text{start}}$ preserving initial signal integrity and $\beta_{\text{end}}$ ensuring complete noise injection by the final step. 

For ensemble forecasting, we implement $P$ independent sampling processes for each input ensemble member $\mathbf{X}^m$. This generates $P \times M$ output ensemble members for an input ensemble with $M$ members. In our implementation, we set $P=20$ to maintain a balance between ensemble diversity and computational efficiency. 

This iterative denoising approach preserves spatial correlation structures through progressive refinement, addressing the spatial independence limitation identified in \citet{tian2025improving} through different mechanisms than QNN or VNN. The resulting predictions maintain both meteorological ensemble diversity and spatially coherent uncertainty structures. 

These probabilistic models quantify regression uncertainty through different mechanisms. QNN directly characterises conditional distribution shapes by predicting multiple quantiles simultaneously. VNN generates diverse outputs through latent space sampling. DNN learns complex conditional distributions through iterative noising and denoising processes. All approaches consider spatial correlations in uncertainty quantification and address the under-dispersion issues in ensemble forecasting. 

\section{Data and metrics}\label{secData}


\subsection{Data} 
To maintain consistency with \citet{tian2025improving}, we use the identical data, \added[id=Ganglin]{partitioning strategy, resolution choice, }preprocessing methods, and study domains. Input variable (Z500) covers the Europe-Atlantic domain (20$^\circ$–80$^\circ$N, 120$^\circ$W–40$^\circ$E), while the target variable (wind speed at 100 meters, U100) focuses on Europe (34$^\circ$–74$^\circ$N, 13$^\circ$W–40$^\circ$E) during the boreal winter\added[id=Ganglin]{, when Z500 anomalies and circulation-wind correlations are strongest \citep{laurila2021climatology}}. \added[id=Ganglin]{The 100-m height is used as it approximates wind turbine hub heights, making it more relevant for wind energy forecasting than 10-m wind. }Both \added[id=Ganglin]{variables }are processed to 2.7$^\circ$ spatial resolution and weekly temporal resolution. Wind speed is derived from its components as $\text{U100}=\sqrt{u^2+v^2}$. 

For the training and validation on reanalysis, we use ERA5 reanalysis data from December 1979 to March 2022, obtained from the Climate Data Store. For evaluation on ensembles, we employ 10-member ECMWF extended-range hindcasts with lead times up to 6 weeks, retrieved from the Meteorological Archival and Retrieval System (MARS), covering December 2015 to March 2021 due to licensing constraints. 

\deleted[id=Ganglin]{Data preprocessing procedures, partitioning strategy, seasonal focus rationale, and resolution choice justification have been thoroughly described in 
, and the same approaches are applied in this study.} 

\subsection{Verification metrics}\label{subsec:verification}
Building on our previous work \citep{tian2025improving}, we assess model performance using both deterministic and probabilistic metrics. Mean Squared Error (MSE) of ensemble mean quantifies deterministic skill, and Continuous Ranked Probability Score (CRPS) and Spread Skill Ratio (SSR) evaluate probabilistic skill. 

\deleted[id=Ganglin]{To quantify the relative improvement of our models over a baseline, we define the relative difference in scores as:
$\Delta_r \text{Score} = \frac{\text{Score}(model_s) - \text{Score}(model_b)}{\text{Score}(model_b)} \times 100 (\%)$
where $model_s$ represents a statistical model being evaluated and $model_b$ represents the baseline model. For negatively oriented scores like MSE and CRPS, negative values of $\Delta_r \text{Score}$ indicate improvement. }

However, these grid-based metrics fail to capture spatial structure characteristics. Therefore, we introduce complementary spatial analysis techniques, including Empirical Orthogonal Function (EOF) decomposition and energy spectrum analysis. \deleted[id=Ganglin]{These additional metrics provide insights into how regression systems represent atmospheric circulation modes, energy distribution across spatial scales, and spatial correlations. }As MSE\added[id=Ganglin]{, CRPS} and SSR calculations were detailed in \citet{tian2025improving}, we focus here on \replaced[id=Ganglin]{EOF analysis and spectrum analysis}{CRPS decomposition, EOF analysis, and spectrum analysis}. 

\deleted[id=Ganglin]{1) CRPS DECOMPOSITION}

\deleted[id=Ganglin]{CRPS quantifies the integrated squared difference between the cumulative distribution function of a probabilistic forecast and that of the corresponding observation, and can be decomposed into three physically meaningful components 
: 
$\text{CRPS} = \text{Reliability} - \text{Resolution} + \text{Uncertainty}$
Reliability quantifies the consistency between forecast probability distributions and observed frequencies of the event, measuring the calibration of the forecast. Resolution, in the context of this specific CRPS decomposition, measures the ability of the forecast to discriminate between different observed situations compared to the climatological forecast. Uncertainty reflects the inherent variability of the predictand. 
Since Uncertainty depends on climatology and remains constant across models, we focus our analysis on Reliability and Resolution. }

\subsubsection{Empirical Orthogonal Function Analysis}\label{metric:EOF}

EOF analysis, also known as principal component analysis in other fields, is a widely used technique for identifying the dominant spatial patterns of variability in meteorological fields \citep{skittides2014wind, qin2000determining, jiang2020principal}. \added[id=Ganglin]{EOF modes are ranked strictly by the percentage of variance they explain, not by a direct correspondence to physical spatial scales. The leading (low-order) EOFs capture the dominant patterns of variability, which are often large-scale circulation patterns because such phenomena typically account for the most variance. Conversely, higher-order EOFs represent more detailed and localized spatial structures that contribute progressively less to the total variance. Therefore, this approach evaluates the field's structure through its different modes of variability, which is different from the scale-dependent verification based on physical wavelengths that we address through energy spectrum analysis (next section). }For a predictand field $\mathbf{Y} \in \mathbb{R}^{N \times G}$ from ERA5 reanalysis, EOF decomposition yields: 

\begin{equation}
    \mathbf{Y} = \sum_{k=1}^{K} \text{PC}_{k} \cdot \mathbf{\text{EOF}}_{k}
\end{equation}

where $\text{PC}_{k} \in \mathbb{R}^{N}$ represents the temporal coefficients (principal components) of the $k$-th \added[id=Ganglin]{EOF }mode \added[id=Ganglin]{$\text{EOF}_{k} \in \mathbb{R}^{G}$}\deleted[id=Ganglin]{at initialisation date $n$ and $\text{EOF}_{k} \in \mathbb{R}^{G}$ denotes its corresponding spatial field}. $N$ represents the number of initialisation dates, $G$ is the total number of grid points in the European domain, and $K$ is the maximum number of \added[id=Ganglin]{EOF }modes (equal to $G$ in principle). \deleted[id=Ganglin]{These EOFs are ordered by the amount of variance they explain, with lower-order modes typically capturing the large-scale, physically meaningful patterns in the data. }

Our verification framework uses this EOF decomposition in three sequential steps. First, we compute the EOFs from the ERA5 predictand reanalysis fields to establish the reference spatial patterns. Next, we project both the ECMWF ensemble hindcasts and our model outputs onto these reference EOFs to obtain their corresponding principal components $\hat{\text{PC}}_{k}$. Finally, we reconstruct the regressed fields $\hat{\mathbf{Y}}^{(K')}$ using an increasing number of modes ($K'$): 
\begin{equation}
\hat{\mathbf{Y}}^{(K')} = \sum_{k=1}^{K'} \hat{\text{PC}}_{k} \cdot \mathbf{\text{EOF}}_{k}
\end{equation}

For each $K'$-mode reconstruction, \replaced[id=Ganglin]{we calculate relative skill scores for both MSE and CRPS metrics, specifically the Mean Squared Skill Score (MSSS) and Continuous Ranked Probability Skill Score (CRPSS), with detailed formulations provided in Appendix \ref{appendix:skill_score}.}{we calculate standard verification metrics (MSE, CRPS, SSR) to evaluate performance across different spatial scales.} \deleted[id=Ganglin]{This scale-dependent approach reveals whether models excel at capturing large-scale patterns (lower modes) or smaller-scale features (higher modes), providing crucial insights for applications where accurate representation of spatial structure is essential. }

\subsubsection{Energy Spectrum Analysis}\label{metric:spectrum}

Energy spectrum analysis provides a complementary perspective on spatial verification by quantifying the distribution of variance across different spatial wavelengths \citep{price2025probabilistic,rasp2023weatherbench,lam2022graphcast}. This approach is particularly useful for identifying systematic biases in the representation of spatial scales, such as excessive smoothing (under-representation of small scales) or noise (over-representation of small scales). For a latitudinal predictand field $\mathbf{Y}_{i} \in \mathbb{R}^{N \times G_{lon}}$, we compute the zonal energy spectra along the line of constant latitude at latitude index $i$: 
\begin{equation}
S_{\mathbf{Y}_{i}}(k_z) = \begin{cases}
C_{\phi}|F_{\mathbf{Y}_{i}}(k_z)|^2, & k_z = 0 \\
2C_{\phi}|F_{\mathbf{Y}_{i}}(k_z)|^2, & k_z = 1,2,\ldots,\lfloor G_{lon}/2 \rfloor
\end{cases}
\end{equation}

where $F_{\mathbf{Y}_{i}}(k_z) = \frac{1}{G_{lon}}\sum_{j=0}^{G_{lon}-1} \mathbf{Y}_{i}e^{-i2\pi k_z j/G_{lon}} $, $F_{\mathbf{Y}_{i}}(k_z)\in \mathbb{R}^{N}$ is the Fourier transform coefficient for the latitude band, and $G_{lon}$ is the number of longitudinal grid points, $C_{\phi} = C_0\cos(\phi_i) \in \mathbb{R}$ with $C_0$ being the equatorial circumference (equal to $2\pi \times$ Earth radius), and $\phi_i$ represents the latitude at index $i$. The factor of 2 accounts for the symmetric contributions from positive and negative frequencies in the real-valued field. 

Following \citet{rasp2023weatherbench}, we average the energy spectra across latitudes to obtain a representative energy spectrum for the European domain, accounting for the latitude-dependent Earth surface area. 

\begin{equation}
    S_{\mathbf{Y}}(k_z) = \frac{1}{G_{lat}} \sum_{i}^{G_{lat}} S_{\mathbf{Y}_{i}}(k_z)
\end{equation} 
where $G_{lat}$ is the number of latitude bands. The energy spectra $S_{\mathbf{Y}}(k_z) \in \mathbb{R}^{N \times G_{lat} }$ provides a measure of the variance in field $\mathbf{Y}$ at each spatial scale corresponding to wavenumber $k_z \in \{0,1,2,...,\lfloor G_{lon}/2 \rfloor\}$. For our regional European domain, each wavenumber $k_z$ corresponds to a physical wavelength calculated as the inverse of the spatial frequency. This frequency is determined by the wavenumber divided by the longitudinal spacing \deleted[id=Ganglin]{(in metres) }at each latitude, accounting for the spherical geometry of the Earth. This approach ensures accurate wavelength estimation within our limited spatial domain rather than assuming a global circumference. 

By comparing the energy spectra of regressed outputs to those of ERA5 reanalysis, we can systematically evaluate model performance across spatial scales and identify potential biases in the representation of spatial variability. To facilitate comparison, we compute the \replaced[id=Ganglin]{Relative Energy Spectra Skill score (RESS)}{relative energy spectra skill} score as:

\begin{equation}
\text{RESS}(k_z) = 1 - \frac{S_{\hat{\mathbf{Y}}}(k_z)}{S_{\mathbf{Y}}(k_z)}
\label{eq:ress}
\end{equation}\deleted[id=Ganglin]{,} 

where $S_{\hat{\mathbf{Y}}}(k_z)$ is the energy spectra of model outputs $\hat{\mathbf{Y}}$ and $S_{\mathbf{Y}}(k_z)$ is the energy spectra of reference (ERA5 reanalysis $\mathbf{Y}$). This \replaced[id=Ganglin]{relative energy spectrum skill}{Relative Energy Spectra Skill score (RESS)} provides a normalised measure of spectral bias. Values close to 0 indicate perfect agreement between model and reference \replaced[id=Ganglin]{spectrum}{spectra}. Positive RESS values indicate energy under-estimation by the model at wavenumber $k_z$, while negative values indicate energy over-estimation. The magnitude of RESS reflects the degree of bias, with larger absolute values signifying greater discrepancy from the reference \replaced[id=Ganglin]{spectrum}{spectra}.

\subsubsection{\added[id=Ganglin]{Statistical significance testing}}

\added[id=Ganglin]{To ensure the robustness of our results, we employ bootstrap resampling for statistical significance assessment. Following \citet{tian2025improving}, we use bootstrap techniques similar to \citet{goddard2013verification}. For each comparison, we randomly select samples with replacement from model outputs to generate 1000 bootstrap replicates, maintaining the same sample size as the original data. }

\added[id=Ganglin]{We calculate verification scores for each bootstrap sample and compute the relative difference $\Delta_r \text{Score} = \frac{\text{Score}(model_s) - \text{Score}(model_b)}{\text{Score}(model_b)} \times 100 (\%)$ between models. For negatively oriented scores (MSE, CRPS), the proportion of $\Delta_r \text{Score}<0$ serves as the $p$-value, while for positively oriented scores, the proportion of $\Delta_r \text{Score}>0$ provides the $p$-value. All line plots display median values from bootstrap distributions, while spatial maps indicate statistical significance with different markers. 
}


\section{\replaced[id=Ganglin]{Results}{Results on hindcasts}}\label{secResults_HC}
\replaced[id=Ganglin]{This section evaluates whether grid-wise metrics adequately capture model performance and how different uncertainty quantification methods preserve spatial coherence. The ECMWF ensemble hindcasts are first calibrated using Mean-Variance Adjustment \citep{goutham2022skillful,goutham2023statistical,manzanas2019bias,tian2025improving} to remove systematic bias and adjust ensemble spread, serving as our benchmark. This calibration enables the statistical models to focus on capturing the predictable Z500-U100 relationship. The evaluation proceeds in four parts: spatially averaged scores across lead times (weeks 1-6), spatial distribution of improvements at week 3, EOF analysis, and energy spectrum analysis.}{The performance evaluation on reanalysis only reflects model behaviour under ``perfect prediction'' conditions. In practical forecasting applications, models must deal with imperfect inputs from weather forecast systems. This section investigates the skill of these models in sub-seasonal hindcasts using ECMWF as a benchmark. 
We first analyse the spatially averaged scores of each model across different lead times. Subsequently, we focus on the third forecast week, conducting analyses through spatial distribution maps ($\Delta_r$MSE and $\Delta_r$CRPS), EOF reconstruction analysis, and zonal relative spectrum skill to explore skill differences on the sub-seasonal time scale.} 

\subsection{Spatially averaged scores}

\begin{figure*}[htb!]
    \centering
    \includegraphics[width=\linewidth]{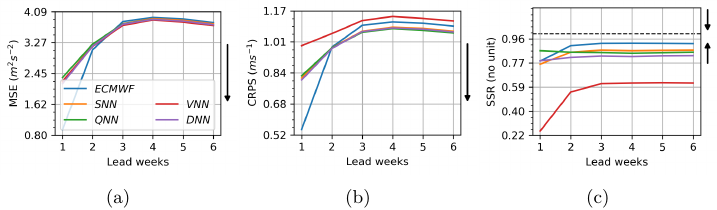}
    \caption{Spatially averaged MSE, CRPS and SSR of different statistical downscaling models and ECMWF ensemble hindcasts across lead weeks. \added[id=Ganglin]{Black arrows indicate optimal directions: MSE and CRPS (lower is better), SSR (closer to 1, shown as black dashed line, is better).}}\label{fig:CurveHC}
\end{figure*}

Figure \ref{fig:CurveHC} illustrates the spatially averaged MSE, CRPS and SSR of different models across lead weeks. In the first week, all statistical downscaling models exhibit relatively low MSE\deleted[id=Ganglin]{ (2.16-2.32 $(m^2s^{-2})$)}, as shown in Figure \ref{fig:CurveHC}a, but they are significantly higher than ECMWF\deleted[id=Ganglin]{ (0.94 $(m^2s^{-2})$)}. 
\deleted[id=Ganglin]{As the forecast lead time extends, the MSE of these models increases rapidly, reaching 3.16-3.21 $(m^2s^{-2})$ by week 2}. From week 3 onwards, the MSE of statistical models begins to approach or even slightly exceed ECMWF. \replaced[id=Ganglin]{This convergence reflects the diminishing benefit of initial conditions in ECMWF hindcasts at longer lead times, where predictability increasingly depends on Z500-U100 relationships that statistical models can effectively capture. }{This indicates that although these statistical models perform less favourably than ECMWF in the first weeks, they exhibit comparable or better MSE after week 3. }

In Figure \ref{fig:CurveHC}b, CRPS results reveal more pronounced differences. \replaced[id=Ganglin]{During weeks 1-2, SNN, QNN, and DNN exhibit similar CRPS performance to their MSE behaviour, underperforming ECMWF. However, from week 3 onwards, these three models achieve better CRPS than ECMWF. In contrast, VNN consistently underperforms ECMWF throughout all lead times.}{In the first week, SNN, QNN, and DNN perform similarly, significantly outperforming VNN but still underperforming compared to ECMWF. As the lead week extends, differences among these models gradually diminish, with QNN achieving the optimal CRPS (1.05 $(ms^{-1})$) by week 3, superior to ECMWF (1.09 $(ms^{-1})$) and other models (1.06-1.12 $(ms^{-1})$). From week 3 onwards, SNN, QNN and DNN maintain CRPS at 1.06-1.08 $(ms^{-1})$, still better than ECMWF, indicating their ability to handle the growing ensemble uncertainty over time.} 

\replaced[id=Ganglin]{To understand these CRPS differences, we examine the SSR in Figure \ref{fig:CurveHC}c. VNN shows substantially lower SSR than other models across all lead times, indicating under-dispersed ensembles that fail to adequately represent forecast uncertainty. While SNN, QNN, and DNN also exhibit under-dispersion (SSR $<$ 1), they remain much closer to the ideal SSR than VNN.}{The SSR (Figure \ref{fig:CurveHC}c) reveals the evolution of ensemble dispersion for each model. QNN and SNN already demonstrate near-ideal SSR (0.87 and 0.77) in week 1, comparable to ECMWF (0.79). DNN's SSR remains stable at 0.79-0.83. In contrast, VNN exhibits significantly lower SSR (0.25) in week 1, with improvements but still insufficient (0.62 by week 3). This suggests that VNN generates under-dispersed ensembles, failing to represent ensemble uncertainty adequately. As the lead week extends, ECMWF's SSR rises to 0.93, slightly exceeding all the statistical models, indicating that its calibrated ensemble members better represent ensemble uncertainty in longer lead weeks.} 

\added[id=Ganglin]{These results indicate that, except for VNN, the statistical models achieve comparable spatially averaged MSE and CRPS performance. VNN's under-dispersion, despite its reasonable mean-state accuracy, suggests biases in its probabilistic representation. This may be attributed to the constraints imposed by the normal distribution assumption in the latent space, which might not fully capture the complex uncertainty structure inherent in wind speed fields.}

\subsection{Spatial distribution of improvements}

\begin{figure*}[htb!]
    
    \centering
    \includegraphics[width=\linewidth]{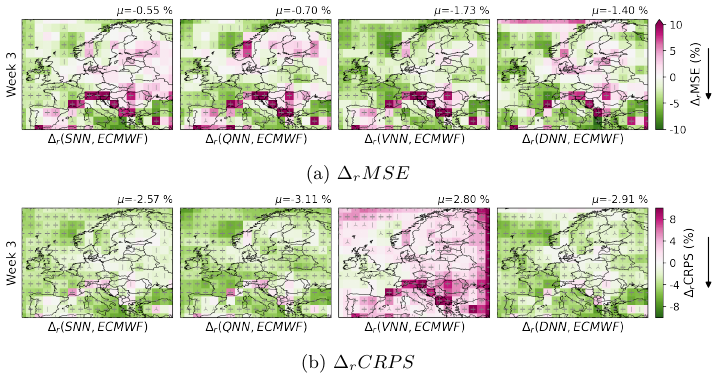}
    \caption{
        \replaced[id=Ganglin]{
            Spatial distribution of the relative differences in MSE and CRPS ($\Delta_r$ MSE and $\Delta_r$ CRPS) between SNN, QNN, VNN and DNN compared to Climatology. Green grids indicate improvements (negative values), while pink grids indicate degradation (positive values). Statistical significance is assessed using bootstrap resampling and indicated by symbols overlaid on grid points: ``+'' denotes significance at $\alpha$ = 0.01, ``Y'' at $\alpha$ = 0.05, and ``I'' at $\alpha$ = 0.1. The $\mu$ value at the top of each subfigure represents the average relative percentage difference across Europe.
        }{
            Same as \protect{\ref{fig:DeltaMapsRA}}, but for hindcasts at lead week 3.
        }    
    }
    \label{fig:DeltaMapsHC}
\end{figure*}

Figure \ref{fig:DeltaMapsHC} illustrates the spatial distribution of relative improvements in \replaced[id=Ganglin]{MSE and CRPS}{CRPS and MSE} for different models compared to \added[id=Ganglin]{calibrated }ECMWF at week 3. 
\replaced[id=Ganglin]{The spatially averaged MSE improvements are modest, showing the similar performance across models observed in Figure \ref{fig:CurveHC}. This modest improvement also reflects the limited room for improvement left by ECMWF calibration, which has already removed systematic bias. In contrast, statistical models achieve more substantial CRPS improvements relative to calibrated ECMWF, demonstrating their advantage in uncertainty quantification.}{Unlike the improvements of approximately 45-47\% observed on reanalysis (Figure \ref{fig:DeltaMapsRA}a), the average improvement on hindcasts ranges from only 0.55-1.73\%, with substantial regional variations. This significant reduction indicates that these statistical models provide limited additional deterministic information when input fields already contain uncertainties, likely also reflecting the high quality of ECMWF hindcasts that leaves minimal room for further improvements. Unlike the improvements of approximately 30\% observed across models on reanalysis (Figure \ref{fig:DeltaMapsHC}b), the average improvements of CRPS with ensemble forecasting are reduced. SNN, QNN, and DNN achieve only 2.57\%, 3.11\%, and 2.91\% improvements, respectively, compared to ECMWF. This suggests a notable increase in the challenge of representing uncertainty on hindcasts. While VNN performed adequately on reanalysis, its overall performance deteriorates on hindcasts (average improvement +2.80\%), indicating sensitivity to input uncertainties.} 

\replaced[id=Ganglin]{Both MSE and CRPS improvements exhibit similar spatial patterns. Improvements are concentrated in Northern and Western Europe, while performance remains limited over mountainous regions such as the Alps and Balkan Peninsula. This highlights the challenge of representing ensemble uncertainty in areas with complex topography, where topographic influences on wind fields occur at scales finer than those represented in the Z500 predictor field. }{Significant changes are also observed in spatial distribution patterns. Whereas the improvements are concentrated over the North European Plain on reanalysis, they shift towards Northern and Western Europe on hindcasts. This shift in spatial pattern is primarily due to the different reference models used in each comparison, climatology (statistical) in Section \ref{secResults_RA} versus ECMWF (dynamical) here. The ECMWF model likely has varying skill across different regions, performing better in some areas than others, which affects where our statistical models can provide the most improvement. Of particular concern is the consistently poor performance of these models over the Alps mountainous region and the Balkan Peninsula, where their results significantly underperform compared to reanalysis outcomes. This highlights the difficulty of ensemble uncertainties in areas with complex topography. }

\subsection{Skill of reconstructed field\added[id=Ganglin]{s} from EOF analysis}
\begin{figure*}[htb!]
    
    \centering
    \includegraphics[width=\linewidth]{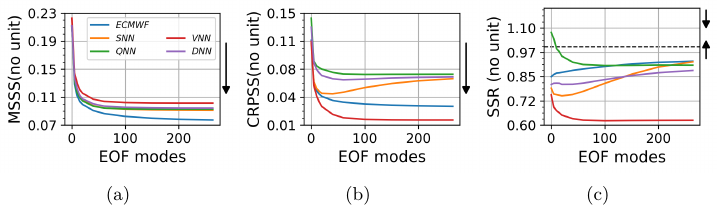}
    \caption{
    \replaced[id=Ganglin]{
    Skill scores of reconstructed fields by EOF modes for different models on hindcasts. (a) MSSS of ensemble mean, (b) CRPSS, and (c) SSR vary with the number of EOF modes used. Black arrows indicate optimal directions (as in Figure \ref{fig:CurveHC}): MSSS and CRPSS (higher is better); SSR (closer to 1, shown as black dashed line, is better).}{Same as \protect{\ref{fig:EOFRA}}, but for hindcasts at lead week 3, using ECMWF as a benchmark.}}
    
    \label{fig:EOFHC}
\end{figure*}

\replaced[id=Ganglin]{Figure \ref{fig:EOFHC} assesses model performance in reproducing spatial structures at week 3, based on field reconstructions using a growing number of EOF modes. Using only the first few modes isolates the dominant and high-variance patterns, while including more modes incorporates finer spatial details. 
}{Figure \ref{fig:EOFHC} presents the scores of various models in reconstructing fields using different numbers of principal components at week 3, compared to those from reanalysis (Figure \ref{fig:EOFRA}). MSSS drops from a range of 0.50-0.78 on reanalysis to 0.08-0.22 on hindcasts, representing a reduction of approximately 60-70\%. Similarly, CRPSS experiences a decline, from 0.27-0.49 on reanalysis to 0.01-0.14 on hindcasts. This reduction in skill can be attributed to the use of ECMWF Z500 hindcasts as inputs rather than the more accurate ERA5 reanalysis, indicating the challenge of maintaining skill when working with imperfect input fields. Interestingly, the introduction of forecast uncertainty led to an improvement in VNN's dispersion characteristics. }

\deleted[id=Ganglin]{Results on hindcasts reveal the skill of ECMWF. Its MSSS are comparable to those of the statistical models at low principal component modes but slightly lower at higher modes. ECMWF's SSR increases gradually from 0.85 to 0.93 and even surpasses the skill scores of all the statistical models at high principal component modes. }


\replaced[id=Ganglin]{When reconstructing these dominant patterns, all models perform similarly against the ECMWF benchmark, achieving high MSSS and CRPSS scores. Skill scores for most models converge around the 20th EOF mode, indicating a common limit to the predictability of fine-scale details at this forecast lead time.}{The MSSS and CRPSS convergence patterns shift earlier on hindcasts, typically appearing around the 20th component when the statistical models begin to show convergence. This earlier convergence suggests that at week 3 lead time, the predictability of smaller-scale features is substantially limited. After a certain number of principal components, adding more spatial details does not improve forecast skill because these finer-scale features cannot be accurately predicted at this forecast range, regardless of the model used, except SNN.} 
\replaced[id=Ganglin]{
SNN is an exception. It underperforms in capturing the dispersion of the primary large-scale patterns, as shown by its lower SSR at low mode counts. Yet, its CRPSS improves as more modes are added, rising from 0.04 to 0.06 between the 40th and 266th modes. This is because SNN’s grid-independent perturbations introduce artificial noise, which primarily inflates the variance in higher-order EOFs. Consequently, SNN creates a favourable ensemble spread, but this improvement is an artifact of non-physical noise and it masks the model's failure to represent large-scale circulation.
}
    {SNN's CRPSS pattern, following a ``down-and-up'' pattern, is also evident on hindcasts, with values rising from 0.04 at the 40th principal component to 0.06 at the 266th principal component. 
    In contrast, the relative amplitude of this change on reanalysis environments is only about 7\%. Notably, SNN's SSR begins at 0.79 and ends at 0.92, both significantly higher than its SSR on reanalysis (0.16 and 0.78, respectively). These pronounced differences in SNN's behaviour between reanalysis and hindcasts stem from the combination of two uncertainty sources, including the inherent uncertainty in the Z500 ensembles from dynamical forecasts and the stochastic perturbations applied during regression. On reanalysis, only the regression uncertainty (stochastic perturbations) is present, while on hindcasts, both sources contribute to the total uncertainty, resulting in the observed differences in spread characteristics. }

\subsection{Energy Spectrum Analysis on Hindcasts}

\begin{figure*}[htb!]
    \centering
    \includegraphics[width=\linewidth]{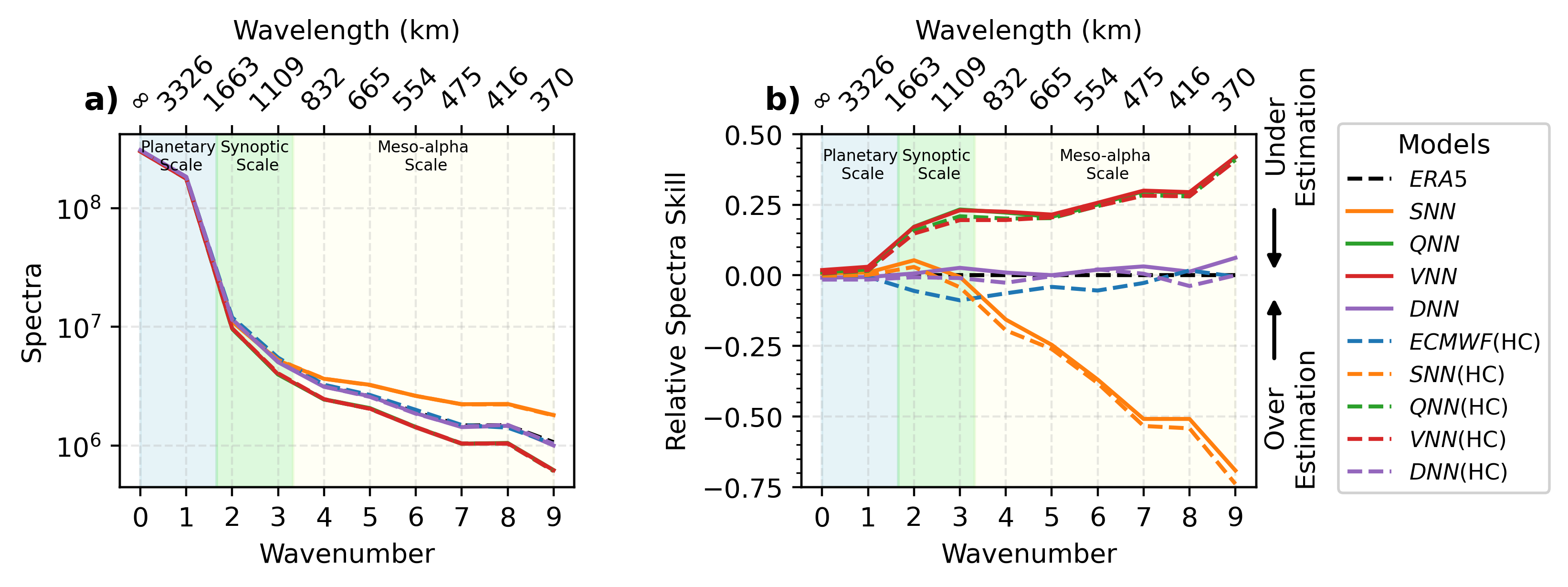}
    \caption{Energy spectrum analysis on hindcasts at lead week 3 compared with reanalysis data. (a) Absolute energy spectra displayed on a logarithmic scale showing similar patterns between solid lines (reanalysis results) and dashed lines (hindcast results) for corresponding models. (b) Relative energy spectrum skill compared to ERA5 reference, where positive values indicate energy under-estimation and negative values indicate over-estimation. \added[id=Ganglin]{The black arrow and black dashed line at zero indicate the optimal direction: RESS values closer to zero represent better spectral reproduction.} The horizontal axis represents zonal wavenumbers with corresponding wavelengths (km). \deleted[id=Ganglin]{While the absolute spectra (a) show minimal visible differences between reanalysis and hindcast environments, the relative skill metric (b) reveals that model-specific characteristics persist but with modified magnitudes in the forecast environment.}}
    \label{fig:SpectrumHC}
\end{figure*}

Figure \ref{fig:SpectrumHC} \replaced[id=Ganglin]{compares the spectra of reanalysis (solid lines) and hindcast (dashed lines) for downscaling, further examining spatial coherence preservation under forecast uncertainty.}{presents a comprehensive comparison between reanalysis (solid lines) and hindcast (dashed lines) spectral characteristics, revealing how forecast uncertainty interacts with each model's inherent spatial representation capabilities.} \added[id=Ganglin]{Note that we use linear wavenumber scaling rather than the conventional log-log representation due to our limited spectral range at 2.7$^\circ$ resolution.} The absolute energy spectra (Figure \ref{fig:SpectrumHC}a) demonstrate \replaced[id=Ganglin]{consistency}{stability} between reanalysis and hindcast environments for the models. \deleted[id=Ganglin]{The logarithmic scale reveals that the fundamental energy cascade remains intact, with the models maintaining the characteristic power-law decay from large to small scales. }\replaced[id=Ganglin]{The agreement between reanalysis and hindcast energy spectra confirms the robustness of statistical downscaling methods to forecast uncertainty in the input fields. Nevertheless, this spectral consistency obscures the modest inter-model differences that emerge through relative skill assessment.}{This stability suggests that the basic physical structure of the fields is preserved through the statistical downscaling process, even when applied to uncertain inputs. However, this apparent similarity masks important differences that become evident only through relative skill analysis.} 

The \replaced[id=Ganglin]{RESS}{relative energy spectrum skill} comparison (Figure \ref{fig:SpectrumHC}b) reveals \replaced[id=Ganglin]{model-specific adaptations to forecast uncertainty. ECMWF hindcast shows notable energy over-estimation in the mesoscale range.}{how each model's characteristic behaviour adapts to the forecast environment. ECMWF hindcast is introduced as an additional reference, which shows notable energy over-estimation in the mesoscale range (relative skill of -0.06 to -0.09), representing a distinct dynamical model signature not present in those statistical models.} 
\replaced[id=Ganglin]{DNN maintains exceptional multi-scale consistency (RESS$\approx 0$ across all wavelengths), demonstrating that its diffusion process preserves physical energy distribution even with imperfect inputs. }{Examining the quantitative changes from reanalysis to hindcast environments reveals model-specific adaptations. DNN demonstrates exceptional robustness, with relative skill values changing by less than 0.04 across all scales when transitioning from reanalysis to hindcasts. At planetary scales, DNN maintains near-perfect energy matching (-0.02 to -0.01), while at mesoscales, it shows slight improvement from reanalysis (0.06) to hindcasts (0.00). This stability suggests that the diffusion process naturally accommodates forecast uncertainty without compromising its multi-scale energy representation. }
\replaced[id=Ganglin]{SNN's RESS exhibits a pattern of increase followed by decrease, indicating energy under-estimation at large scales (producing overly smooth fields) and over-estimation at small scales (producing excessively noisy fields).
}{SNN exhibits amplified energy over-estimation at small scales in the hindcast environment, with relative skill deteriorating from -0.69 to -0.74 at the smallest wavelengths. This 7\% amplification confirms that forecast uncertainty exacerbates the limitation of spatial independence in the stochastic perturbation approach. The uncorrelated perturbations, when combined with ensemble forecast spread, generate even more excessive small-scale variability that may compromise physical realism in the forecast contexts.} 
\replaced[id=Ganglin]{QNN and VNN show nearly identical patterns in both energy spectra and RESS, consistently under-estimating energy and producing overly smooth wind speed estimates that miss fine-scale variability in wind patterns. The consistency of their RESS between reanalysis and hindcasts indicates that the}{QNN and VNN show remarkable consistency between environments, with changes in relative skill typically less than 0.02 across all scales. Both models maintain their characteristic energy under-estimation at small scales (0.41 for QNN and VNN in hindcasts versus 0.42 in reanalysis). This stability suggests that their} tendency to produce overly smooth fields is an inherent characteristic of their uncertainty representation mechanisms\added[id=Ganglin]{,} rather than a response to input uncertainty.
\deleted[id=Ganglin]{The comparison with ECMWF hindcasts provides additional insights. While ECMWF shows excellent performance at planetary scales, its mesoscale energy over-estimation contrasts with the statistical models' behaviours. This difference highlights a distinction that dynamical models may struggle with appropriate energy distribution at intermediate scales due to parameterisation limitations, while the statistical models' spectral characteristics are primarily determined by their uncertainty quantification mechanisms. }

\replaced[id=Ganglin]{These results complement the EOF analysis by demonstrating that SNN's apparent skill improvement through stochastic perturbations is artificial and physically unrealistic. They also confirm that DNN provides robust spectral representation under forecast uncertainty, while other models exhibit characteristic limitations that persist or amplify when applied to uncertain inputs.}{These spectral characteristics in both reanalysis and hindcast environments complement and extend the findings from 
. While 
identified the issue of spatial independence in the stochastic perturbation approach, the spectrum analysis here quantifies exactly how this independence manifests as excessive small-scale energy, and how this limitation is further amplified in forecast environments. The analysis reveals that more sophisticated approaches like DNN address this limitation through better spectral matching that remains robust even under forecast uncertainty. }

\section{\replaced[id=Ganglin]{Discussion and Conclusions}{Discussion}}\label{secDiscussion}
\replaced[id=Ganglin]{This study addressed two questions about uncertainty quantification in sub-seasonal wind speed forecasting: (1) whether grid-wise metrics adequately evaluate probabilistic downscaling methods, and (2) how different uncertainty quantification approaches preserve spatial correlation structures. By comparing stochastic perturbations, quantile regression, variational autoencoders, and diffusion models on ECMWF sub-seasonal hindcasts, we demonstrate the necessity of spatial uncertainty assessment to complement grid-wise evaluation.}{This study compares the performance of different uncertainty representation methods in sub-seasonal wind speed prediction, conducting a comprehensive evaluation from reanalysis data to operational hindcasts. The following sections examine our results from several theoretical perspectives, including model mechanisms, spatial characteristics, and forecast evolution. }

\subsection{\added[id=Ganglin]{Uncertainty Mechanisms and Spatial Consistency}}

\added[id=Ganglin]{All uncertainty quantification methods achieve comparable or superior performance to calibrated ECMWF hindcasts on grid-wise metrics. At week 3, statistical models except VNN show similar MSE with modest CRPS differences, and exhibit lower wind speed skill in complex terrain regions. This apparent skill consistency might suggest that simple stochastic perturbations provide adequate uncertainty quantification for sub-seasonal forecasting. However, grid-wise metrics evaluate only skill at independent grids, failing to capture how uncertainty is spatially structured. This is a limitation for applications where spatial correlations influence aggregate variability, extreme event extent, and regional diversification strategies. }

\added[id=Ganglin]{Spatial analyses reveal fundamental differences tied to each method's theoretical foundation. SNN's grid-independent noise creates unrealistic spatial patterns, evidenced by both excessive small-scale energy in the spectrum and artificially inflated variance in higher-order EOF modes. This artifact masks the model's failure to capture the dispersion of dominant large-scale patterns. QNN and VNN both produce overly smooth spatial structures due to their optimisation objectives. QNN's quantile loss optimises marginal distributions at each grid point without spatial correlation constraints, leading to small-scale energy under-estimation. VNN's under-dispersion stems from two interrelated constraints. The KL divergence regularisation toward Gaussian priors limits latent space expressiveness, while the underlying Gaussian assumption cannot fully capture the non-Gaussian complexity inherent in wind field uncertainty. This manifests as both spatial smoothing and insufficient ensemble spread, despite adequate mean-state representation. DNN's iterative denoising operates on entire spatial fields, preserving multi-scale structures through gradual noise refinement that aligns with physical energy cascade processes, achieving near-zero RESS across all wavelengths. }

\subsection{Methodological limitations and future directions}
\replaced[id=Ganglin]{This study has several limitations that point toward future research directions. First, all statistical models presented here only use Z500 as the predictor variable. While this choice was intentional based on preliminary experiments showing that additional variables such as Z50 and SST provided minimal skill improvement, the potential benefits of multi-variate predictors need further investigation for different regions and seasons. }{Despite providing a systematic evaluation, this study has several limitations. First, all the statistical models presented here rely on Z500 as the predictor variable. While incorporating additional atmospheric variables (e.g., Z50, SST) might seem beneficial, our preliminary experiments indicated that adding these predictors did not significantly improve forecast skill. This finding suggests that Z500 may already capture the most relevant large-scale circulation features for sub-seasonal wind speed prediction in our study region. }
\replaced[id=Ganglin]{Second, the 2.7$^\circ$ spatial resolution balances computational efficiency with meteorological representation while filtering unpredictable noise. However, this coarse resolution limits our ability to resolve fine-scale features in complex terrain. Finer resolution could better capture topographical influences and local circulations, though the computational cost may outweigh marginal skill improvements at sub-seasonal timescales. }{Second, the fixed spatial resolution (2.7$^\circ$) used in this study was chosen to balance computational efficiency with meteorological representation. This coarse resolution effectively filters out unpredictable noise. Nevertheless, it does limit our ability to resolve fine-scale wind field features, particularly in regions with complex terrain. While using finer resolution could potentially benefit sub-seasonal prediction by better resolving topographical influences and local circulation patterns, the trade-off between increased computational demands and marginal improvements in predictive skill at the sub-seasonal timescale must be carefully considered. }
\replaced[id=Ganglin]{Third, our temporal independence assumption treats each lead time separately, ignoring temporal correlations in wind evolution. Sequence modeling approaches incorporating temporal dependencies, such as spatio-temporal convolutional networks or recurrent diffusion architectures, could potentially improve performance by better capturing wind field evolution while maintaining physical consistency. }{Third, our training methodology employs a temporal independence assumption, treating each lead time step separately without considering temporal correlations in wind speed evolution. Sequence modelling approaches that incorporate temporal dependencies could potentially enhance prediction performance, particularly spatio-temporal convolutional networks or recurrent architectures combined with diffusion processes, which might better capture wind field temporal evolution while maintaining physical consistency. }
\replaced[id=Ganglin]{Finally, we compared individual uncertainty methods without exploring hybrid or multi-model combinations that might leverage their complementary strengths. Future work should also evaluate model performance for extreme wind events, which are particularly critical for wind energy risk management. }{Finally, we compared individual uncertainty representation methods without exploring hybrid approaches or multi-model combination approaches that might combine their respective strengths. Additionally, future work should specifically evaluate model performance for extreme wind events, which are particularly relevant for risk management applications in the wind energy sector. }

\subsection{Practical applications and implications}

Our findings have direct implications for wind energy applications and operational meteorological forecasting. DNN's superior performance in physical consistency makes it particularly suitable for applications where accurate representation of wind field physical properties is crucial \citep{pickering2020sub,grams2017balancing}. \deleted[id=Ganglin]{Additionally, DNN demonstrates excellent performance in deterministic skill and probabilistic skill, enabling it to provide comprehensive support for both operational decision-making and risk management. }\replaced[id=Ganglin]{Other models show comparable grid-wise skill, which may suffice for applications that do not require spatial coherence. However, their spatial characteristics impose practical constraints.}{Other models, despite having specific advantages, exhibit notable limitations for practical applications. SNN's small-scale energy over-estimation may lead to over-estimation of local variability and potential extreme events.} VNN's insufficient probabilistic skill limits its applicability for risk assessment applications. \replaced[id=Ganglin]{QNN exhibits small-scale energy under-estimation, potentially degrading performance in local-scale wind forecasting applications. }{QNN, while showing excellent grid-wise skill, exhibits energy under-estimation at small scales, potentially limiting its application for local-scale wind forecasting. }

\added[id=Ganglin]{From a computational perspective, all statistical models require substantially fewer computational resources than dynamical forecasting systems. Inference times range from seconds (SNN, QNN) to several minutes (DNN), with computational complexity increasing approximately in the order $\text{SNN} \approx \text{QNN} < \text{VNN} < \text{DNN}$ due to their respective inference mechanisms. However, all methods remain computationally feasible for operational forecasting applications, making forecast skill the primary consideration in model selection.}

\subsection{Conclusion}\label{secConclusion}

\replaced[id=Ganglin]{This study demonstrates that capturing spatial correlation structures in uncertainty representation is essential for sub-seasonal wind speed forecasting, extending beyond grid-wise skill assessment. Addressing our two research questions, we show that: (1) grid-wise metrics mask critical differences in spatial uncertainty representation, providing incomplete model assessment when spatial coherence matters, (2) among the four uncertainty quantification approaches, only diffusion models consistently preserve multi-scale spatial dependencies, while stochastic perturbations, quantile regression, and variational autoencoders each exhibit distinct spatial limitations. Quantile regression and variational autoencoders produce overly smooth wind fields with under-estimated small-scale energy. Stochastic perturbations over-estimate energy through spatially uncorrelated noise.}{This study evaluates the performance of different uncertainty representation methods in sub-seasonal wind speed prediction, providing new insights into spatial uncertainty representation in statistical downscaling. Through the comparison of stochastic perturbations, quantile regression, variational autoencoders, and diffusion models, we investigated how to effectively represent spatial uncertainty in statistical downscaling while maintaining physical consistency.}

\deleted[id=Ganglin]{Our research reveals important differences among uncertainty representation methods. QNN performs well in probabilistic prediction metrics. DNN shows strengths in physical consistency, with spectrum analysis showing close alignment with ERA5 across all spatial scales. SNN demonstrates advantages in representing small-scale variability, although its small-scale energy over-estimation may lead to over-estimation of local uncertainty. VNN performs well in deterministic prediction but shows notable deficiencies in probabilistic prediction performance and ensemble dispersion. Model performance exhibits clear spatial scale dependence and geographical variation, with the statistical models demonstrating comparable or slightly better forecast skill to ECMWF after week 3. }

The main contributions of this study include the first systematic comparison of the four modern uncertainty representation methods in statistical downscaling and sub-seasonal wind speed ensemble forecasting, proposing a comprehensive evaluation framework combining forecast verification metrics with spatial consistency analysis, and revealing the theoretical connections between uncertainty representation methods and forecast skill. Despite certain methodological limitations such as single predictor use, fixed spatial resolution, and temporal independence assumptions, this study provides \replaced[id=Ganglin]{the important insight that advancing sub-seasonal statistical forecasting requires moving beyond grid-point accuracy toward evaluating spatial uncertainty representation. }{important insights for improving the skill of sub-seasonal wind speed prediction while indicating directions for developing more advanced statistical downscaling methods. }



\acknowledgments
This work has been carried out at the Energy4Climate Interdisciplinary Center (E4C) of IP Paris and Ecole des Ponts ParisTech, which is in part supported by 3rd Programme d’Investissements d’Avenir [ANR-18-EUR-0006-02], and by the Foundation of Ecole polytechnique (Chaire "Défis Technologiques pour une Énergie Responsable” financed by TotalEnergies).

\datastatement
The ERA5 reanalysis data used in this study are available from the Climate Data Store (\url{https://cds.climate.copernicus.eu}). The ECMWF extended-range forecast data were obtained through the Meteorological Archival and Retrieval System with an institutional license. Researchers interested in sub-seasonal forecasts and hindcasts can access related datasets through the ECMWF S2S project (\url{https://www.ecmwf.int/en/research/projects/s2s}). The code used in this study, including the specific configuration of the SmaAt-UNet structure, is available at our GitHub repository: \protect\url{https://github.com/TIANGANGLIN/s2s-spatial-correlation}.

\appendix
\appendixtitle{a}
\subsection{Implementation details}\label{appendix:implementation}

This appendix provides detailed information on the implementation of the probabilistic models described in Section \ref{secMethodology}\ref{subsecModels}, including training framework and hyperparameter optimisation strategies. While the main text focuses on the theoretical foundations and mathematical formulations of each model, here we elaborate on the practical considerations necessary for reproducing our experimental results.

\subsubsection{Training framework}

Our training procedures build upon the training framework established in \citet{tian2025improving}, extending the nested cross-validation framework to accommodate advanced spatial uncertainty quantification. We maintained the same data partitioning, preprocessing steps, and validation strategy while adapting the training process for probabilistic modelling. 

All these models were trained using the Adam optimiser \citep{kingma2014adam} to maintain consistency with \citet{tian2025improving}. However, model-specific learning rate schedules were implemented to accommodate the varying complexity of different probabilistic formulations. Training enhancements implemented in \citet{tian2025improving}, including the Gaussian noise data augmentation strategy ($\sigma$ = 0.1), were maintained to ensure consistency and enable direct performance comparison. 

\subsubsection{Hyperparameter optimisation}

\begin{table}[htb!]
    \centering
    \caption{Model-specific hyperparameter configurations for probabilistic approaches}
    \label{table:hyperparameters}
    \begin{tabular}{ll}
    \hline
    \textbf{Model} & \textbf{Hyperparameters} \\
    \hline
    \multirow{2}{*}{\textbf{QNN}} & Learning rate: $\log\mathcal{U}(10^{-6}, 10^{-1})$ \\
    & Weight decay: $\log\mathcal{U}(10^{-6}, 10^{-1})$ \\
    \hline
    \multirow{4}{*}{\textbf{VNN}} & Learning rate: $\log\mathcal{U}(10^{-6}, 5 \times 10^{-2})$ \\
    & Weight decay: $\log\mathcal{U}(10^{-6}, 10^{-1})$ \\
    & Latent dimension ($d_z$): $\{32, 64, 128, 256, 512, 1024\}$ \\
    & KL weight ($\beta$): $\log\mathcal{U}(10^{-4}, 2 \times 10^3)$ \\
    & Skip connections: With or without for each layer \\
    \hline
    \multirow{7}{*}{\textbf{DNN}} & Learning rate: $\log\mathcal{U}(10^{-6}, 10^{-1})$ \\
    & Weight decay: $\log\mathcal{U}(10^{-6}, 10^{-1})$ \\
    & Noise start ($\beta_{\text{start}}$): $\log\mathcal{U}(10^{-6}, 10^{-3})$ \\
    & Noise end ($\beta_{\text{end}}$): $\log\mathcal{U}(10^{-2}, 2 \times 10^{-1})$ \\
    & Diffusion steps ($T$): $\{100, 200, ..., 1000\}$ \\
    & Time embedding ($d_t$): $\{128, 192, 256, ..., 512\}$ \\
    & DDIM steps: $\{10, 15, 20, ..., 80\}$ \\
    & Randomness: $\{0.0, 0.5, 1.0\}$ \\
    & Temperature: $\{0.5, 1.0, 1.5, 2.0\}$ \\
    \hline
    \end{tabular}
\end{table}

For hyperparameter optimisation, we employed Optuna \citep{optuna_2019} with CRPS as the primary optimisation metric, representing a methodological extension from the MSE-based optimisation in \citet{tian2025improving}. This change reflects the essential capability of our models to generate probabilistic outputs from deterministic inputs: VNN and DNN maintain consistency with \citet{tian2025improving} by generating $P=20$ probabilistic realisations per input Z500 member, while QNN produces 10 quantiles per input member to represent the conditional distribution. The optimisation process involved 400 trials for each model type using Optuna's Tree-structured Parzen Estimator (TPE) sampler, which adaptively focused the search on promising regions of the hyperparameter space, ensuring fair comparison across different uncertainty quantification approaches. Table \ref{table:hyperparameters} summarises the hyperparameter spaces for each probabilistic model. The increasing complexity from QNN to DNN reflects the different theoretical foundations and uncertainty representation mechanisms of each approach.

The QNN architecture requires optimisation of only standard neural network hyperparameters due to its direct quantile prediction approach. As detailed in Table \ref{table:hyperparameters}, QNN's hyperparameter space includes learning rate and weight decay, both sampled from log-uniform distributions ($\log\mathcal{U}$). This relative simplicity compared to other probabilistic methods reflects QNN's straightforward architectural modification of the established CNN framework.

VNN introduces variational-specific hyperparameters beyond standard neural network parameters. The latent dimension $d_z$ controls the information bottleneck critical for balancing reconstruction fidelity with regularisation, while the KL weight $\beta$ implements the $\beta$-VAE framework. Skip connections configuration determines encoder-decoder information flow, with various connectivity patterns explored during hyperparameter optimisation to maintain spatial coherence while enabling effective latent space learning.

DNN exhibits the highest hyperparameter complexity due to its iterative nature. The noise schedule parameters $\beta_{\text{start}}$ and $\beta_{\text{end}}$ define the linear noise scheduling described above, controlling the forward diffusion process. The time embedding dimension $d_t$ provides sufficient capacity for distinguishing different denoising stages in the iterative reverse process. Additional parameters control the trade-off between sampling speed and quality during inference.

\subsection{Skill Scores}\label{appendix:skill_score}

The metrics described in the previous subsections provide absolute measures of forecast performance. However, for model comparison and practical interpretation, it is essential to evaluate performance relative to reference forecasts. Skill scores provide this normalised assessment by expressing model performance as relative improvement over a baseline, typically climatology.

The general form of a skill score is defined as:

\begin{equation}
\text{Skill Score} = 1 - \frac{\text{Score}_{\text{model}}}{\text{Score}_{\text{ref}}}
\end{equation}

where a perfect forecast yields a skill score of 1, no improvement over the reference yields 0, and performance worse than the reference yields negative values. This skill score form applies consistently across the deterministic, probabilistic, and spatial metrics employed in this thesis.

The Mean Squared Skill Score (MSSS) quantifies relative improvement in forecast accuracy:

\begin{equation}
 \text{MSSS} = 1 - \frac{\text{MSE}_{\text{model}}}{\text{MSE}_{\text{ref}}}
\end{equation}

Similarly, the Continuous Ranked Probability Skill Score (CRPSS) evaluates relative improvement in probabilistic forecast quality:
\begin{equation}
 \text{CRPSS} = 1 - \frac{\text{CRPS}_{\text{model}}}{\text{CRPS}_{\text{ref}}}
\end{equation}

\bibliographystyle{ametsocV6}
\bibliography{references}

\end{document}